\documentclass{midl} 

\usepackage{mwe} 

\title[Biopsy Validation]{Cross-Domain Validation of a Resection-Trained Self-Supervised Model on Multicentre Mesothelioma Biopsies}

\midlauthor{
\Name{Farzaneh Seyedshahi\nametag{$^{1,2}$}}
\Name{Francesca Damiola\nametag{$^{3}$}}
\Name{Sylvie Lantuejoul\nametag{$^{3,4}$}}
\Name{Ke Yuan\nametag{$^{1,2}$}}
\Name{John Le Quesne\nametag{$^{1,2,5}$}}\\
\Email{f.seyedshahi.1@research.gla.ac.uk}\\
\addr $^{1}$ School of Cancer Sciences, University of Glasgow, Glasgow, Scotland, UK \\
\addr $^{2}$ Cancer Research UK Scotland Institute, Glasgow, Scotland, UK \\
\addr $^{3}$ Department of Biopathology, Léon Bérard Cancer  Center, Lyon, France\\
\addr $^{4}$ Grenoble Alpes University, Grenoble, France \\
\addr $^{5}$ Queen Elizabeth University Hospital, NHS Greater Glasgow and Clyde, Glasgow, Scotland, UK \\
}

\begin{document}

\maketitle

\begin{abstract}
Accurate subtype classification and outcome prediction in mesothelioma are essential for guiding therapy and predicting patient outcomes. However, most computational pathology models are trained exclusively on large tissue images from resection specimens, which limits their relevance in real-world diagnostic settings where small biopsies are the primary tissue source. Here, we assess the biopsy-level generalisability of a self-supervised encoder using a large, multicentre French cohort. We identify 53 biopsy-specific histomorphological clusters, quantify each patient’s proportional representation across these clusters, and use these profiles as inputs to two downstream tasks: (i) survival prediction using a Cox proportional hazards model and (ii) subtype classification (epithelioid vs. non-epithelioid) using logistic regression. The survival model achieved a test C-index of 0.6 and robustly separated cohort patients into high- and low-risk groups (\(p = 3.96 \times 10^{-29}\)). For subtype classification, the logistic model reached an average AUC of 0.92. These results demonstrate that a self-supervised encoder trained on resection tissue can be reliably transferred to biopsy material despite significant domain shifts. The resulting biopsy-level morphological atlas enables clinically meaningful survival stratification and subtype prediction, supporting the translational integration of AI-driven decision tools in mesothelioma diagnostics.

\end{abstract}


\section{Introduction}

Mesothelioma is a rare but aggressive cancer, most commonly associated with asbestos exposure, and characterised by profound histological heterogeneity and poor prognosis \cite{wagner1960diffuse, molinarimesothelioma}.

Accurate subtype classification, especially into epithelioid, non-epithelioid categories, and reliable survival risk stratification are critical in guiding therapeutic decisions and predicting patient outcomes. However, conventional diagnosis based on histopathology remains subjective and can suffer from high inter-observer variability, particularly in distinguishing transitional morphological patterns \cite{travis20152015, salle2018new, scherpereel2020ers}.

Recent advances in computational pathology have shown promise to address these challenges. For instance, deep-learning models such as MesoNet \cite{courtiol2019deep} have been used to predict overall survival directly from whole-slide images, outperforming standard pathology practice while also uncovering region-level prognostic features rooted in inflammation and stromal architecture. More recently, MesoGraph \cite{eastwood2023mesograph}, a graph neural network (GNN) framework, enabled cell-level scoring of sarcomatoid and epithelioid phenotypes using a small subset of tissue microarrays (TMAs). The method generates a continuous MesoScore that correlates with both subtype composition and patient survival. This work relied on core-level labels within a weakly supervised setup to train the model.

Self-supervised learning has also become an important approach for reducing the high inter-observer variability seen in mesothelioma diagnosis. In contrast to weakly supervised methods that rely on slide/patient/core-level labels, self-supervised models use large unlabeled datasets to learn stable and transferable features, lowering the need for detailed expert annotation. A recent work, Histomorphology Phenotype Learning (HPL) \cite{seyedshahi2025histomorphological}, using a self-supervised learning (SSL) approach trained on resected mesothelioma tissue, has produced a histomorphological atlas that captures recurrent phenotypes across thousands of slides. This atlas has demonstrated clinically meaningful performance for both prognostic stratification and subtype classification. 

Additionally, diagnosis and prognostic assessment in mesothelioma rely mainly on small tissue biopsies, however most computational pathology models are still trained on large tissue samples. Larger resection specimens contain wide and mixed regions of tumour, stroma, necrosis, and inflammation, whereas biopsies are much smaller, more uniform in composition, and more prone to crush artefact, sampling limitations, and staining variability. These differences create a significant mismatch between the data used for model development and the material used in routine diagnostic workflows, with direct implications for model reliability and clinical applicability.

To address this gap, we evaluate a self-supervised model \cite{seyedshahi2025histomorphological} that was originally trained on UK surgical resection WSIs stained with hematoxylin and eosin (H\&E). We apply the trained encoder to a large, multicentre French biopsy cohort stained with HPS (hematoxylin–phloxine–saffron) and HES (hematoxylin–eosin–saffron), thereby introducing substantial differences in both tissue type and staining protocol. This setting introduces two major domain shifts: (a) a change in tissue type and sampling depth (resection to biopsy) and (b) staining differences due to the presence of saffron-based collagen highlighting and phloxine contrast. Additional variations in institution, country, and scanning protocols further increase the clinical relevance of this evaluation.

The key open question is whether self-supervised encoders trained on large resections can be reliably transferred to biopsy material while retaining diagnostic fidelity. Our aim is to address this by evaluating how well the encoder preserves diagnostic and prognostic signals when applied to real-world clinical biopsies. We then assess the utility of biopsy-derived histomorphological clusters for patient-level survival prediction and for accurate classification of epithelioid versus non-epithelioid disease subcategories.

\section{Methods}
\subsection{Datasets}

For this study, we assembled a large multicentre cohort of HES- and HPS-stained biopsy slides collected from pathology laboratories across several regions in France. All slides were centrally reviewed at the Lyon diagnostic centre and digitised using Leica AT2 or Leica GT scanners at 20× magnification. The final dataset includes 976 patients and 1,062 biopsy images. Of these patients, 77\% were male (799) and 23\% female (241). Ages ranged from 22 to 100 years, with a mean of 76.5 and a standard deviation of 8.7. Both age and mesothelioma subtypes were significantly associated with survival (\(p < 0.05\)), with the cohort comprising 657 epithelioid and 405 non-epithelioid cases. All images underwent extensive quality control and preprocessing to ensure sufficient tissue content for downstream inference.

The feature encoder used in this work was trained with the Barlow Twins SSL framework on the Leicester Archival Thoracic Tumour Investigation Cohort–Mesothelioma (LATTICe-M), consisting of 512 patients and 3,446 resection WSIs \cite{seyedshahi2025histomorphological}. For benchmarking, we additionally included the publicly available Cancer Genome Atlas (TCGA)-mesothelioma cohort \cite{hmeljak2018integrative} used previously in the HPL study, which, although substantially smaller (75 patients, 84 images), provides another established external validation set.

\subsection{Model description}
\subsection*{HPL mathematical formulation and validation setup}

The HPL evaluation pipeline consists of three main stages: (i) tile-to-embedding extraction using the Barlow Twins–pretrained encoder, (ii) graph-based Leiden clustering followed by assignment of tiles to clusters, and (iii) construction of patient-level compositional representations with centred log-ratio (CLR) transformation for downstream predictive modelling.

\paragraph{Notation.}
Let \(\mathcal{S}\) denote the set of WSIs in the biopsy cohort. For a WSI \(s\in\mathcal{S}\) we extract a set of non-overlapping tiles:
\[
\mathcal{T}_s = \{t_{s,1}, \dots, t_{s,n_s}\},
\]
where each tile \(t\) is a \(224\times224\) pixles patch at 5\(\times\) (pixel size \(\approx 1.8\ \mu\)m).

\paragraph{Encoder and tile embeddings.}
Let \(f_{\theta}:\mathcal{I}\to\mathbb{R}^D\) be Barlow Twins encoder (Appendix \ref{barlow}) with \(D=128\). For each tile \(t\), its embedding is \(z = f_{\theta}(t)\in\mathbb{R}^D\), optionally \(\ell_2\)-normalised. The encoder was trained on the LATTICe-M dataset, and the resulting pretrained weights were directly used to embed all tiles in the biopsy cohort for inference and evaluation.

\paragraph{HPC discovery and assignation}
A set of histomorphological phenotype clusters (HPCs) \(\mathcal{C}=\{hpc_1,\dots,hpc_c\}\) was derived from the biopsy cohort by constructing a \(k\)-nearest-neighbour graph on a 250{,}000-tile subsample of embeddings and applying Leiden community detection. Given embeddings \(\{z^{(r)}_m\}_{m=1}^{M=250{,}000}\), pairwise distances are computed and each node is connected to its \(k=250\) nearest neighbours, producing an adjacency matrix \(W\). The Leiden algorithm then identifies a partition \(\mathcal{P}\) that maximises modularity with a specific resolution \(\gamma\) (here we picked \(\gamma = 3.0\)). For the rest of the tiles in the cohort, its embedding \(z\) is assigned to the nearest cluster centroid \(\mu_j\) using Euclidean distance:
\begin{equation}
\mathrm{C}(t) = \arg\min_{j\in\{1,\dots,c\}} \| z - \mu_j \|_2.
\label{eq:assign}
\end{equation}
This produces, for each WSI \(s\), cluster counts \(n_{s,j} = |\{t\in\mathcal{T}_s : \mathrm{C}(t)=j\}|\). To ensure a strictly inductive setting, Leiden clustering and the fitting of the cluster centroids were performed exclusively on the feature vectors derived from the training set. The centroids were then fixed and applied to transform the validation features.

\paragraph{Subtype classification (logistic regression).}
For epithelioid vs.\ non-epithelioid classification, we use the frequency of HPCs per WSI to form a slide-level compositional vector. For each slide \(s\), let \(x^{(s)} = \mathrm{clr}(\mathbf{a}^{(s)}) \in \mathbb{R}^c\), where \(\mathbf{a}^{(s)}\) is the HPC frequency vector and \(\mathrm{clr}\) denotes the centred log-ratio transform (Appendix \ref{compositional}). A logistic regression model is then fitted to predict the binary subtype label \(y^{(s)} \in \{0,1\}\).

The predicted probability of the epithelioid subtype is:
\begin{equation}
\Pr\big(y^{(s)} = 1 \,\mid\, x^{(s)}\big)
= \sigma\!\left( \beta_0 + \boldsymbol{\beta}^\top x^{(s)} \right),
\label{eq:logistic}
\end{equation}
where \(\sigma(u) = (1 + e^{-u})^{-1}\). The parameters \((\beta_0, \boldsymbol{\beta})\) are estimated by minimising the regularised negative log-likelihood with an \(\ell_1\) penalty weighted by \(\lambda_{\ell_1}\). \\
We also benchmarked subtype classification using the CLAM (Clustering-constrained Attention Multiple Instance Learning) method \cite{lu2021data}. The method was applied directly to tile embeddings (\(z\)), allowing us to assess performance without any clustering step. Each WSI was treated as a bag of tile embeddings, and the model aggregated instance-level features into a slide-level representation. A linear classifier was attached to the CLAM output to predict epithelioid vs non-epithelioid subtype. CLAM was trained for 30 epochs with early stopping, using Adam (\(lr = 10^{-3}\)) and a binary loss function. The number of attention clusters was fixed at 8. Slide-level subtype labels supervised the model, while tile-level attention scores identified the most informative regions. 

\paragraph{Survival modelling (Cox proportional hazards).}
For patient-level survival analysis, we construct the CLR-transformed HPC composition vector \(x^{(i)}\) for each patient. Let \(T^{(i)}\) denote the observed time and \(\delta^{(i)}\) the event indicator. A Cox proportional hazards model is then fitted:
\begin{equation}
h\big(t \mid x^{(i)}\big) = h_0(t)\exp\big( \gamma^\top x^{(i)} \big),
\label{eq:cox}
\end{equation}
where \(\gamma\) are the log-hazard coefficients. Parameters are estimated by maximising the partial likelihood:
\begin{equation}
\mathcal{L}_{\text{partial}}(\gamma) = \prod_{i:\delta^{(i)}=1}
\frac{\exp(\gamma^\top x^{(i)})}{\sum_{j\in\mathcal{R}(T^{(i)})}\exp(\gamma^\top x^{(j)})},
\label{eq:partial}
\end{equation}
with \(\mathcal{R}(t)\) the risk set at time \(t\). Hazard ratios \(e^{\gamma_j}\) with 95\% confidence intervals are reported. Patients are stratified into high and low-risk groups based on the cohort median risk, and Kaplan–Meier plots illustrate the corresponding survival probabilities.

\paragraph{Performance metrics and evaluation protocol.}
Validation used a patient-wise five-fold cross-validation ensuring strict separation of patients between folds (no tile or WSI from the same patient appears in both training and test splits). For classification we report area under the ROC curve (AUC), accuracy, recall, precision, F1-score and the mean ± standard deviation across folds. For survival, we report Harrell’s concordance index (c-index) with bootstrap 95\% confidence intervals. Where applicable, we test coefficient significance using Wald tests and perform permutation testing (shuffling labels) to evaluate the null distribution of metrics.

\paragraph{Implementation details and reproducibility.}
Embeddings were computed using the publicly available HPL Encoder \cite{adalberto_claudio_quiros_2025_16947385}. We adopted the ResNet backbone as the feature encoder. This encoder was originally trained and validated on the LATTICe-M dataset and subsequently validated on the TCGA mesothelioma cohort in a previous study \cite{seyedshahi2025histomorphological}. In the present work, we followed the same validation strategy: we directly used the trained encoder weights to extract feature representations from biopsy tiles, without performing any additional fine-tuning or retraining (Figure \ref{fig:hpl}a). Although, unlike the TCGA-Meso validation, which relied on a predefined dictionary of histomorphological clusters derived from resection specimens, we re-clustered the feature embeddings from scratch. This approach allowed us to identify biopsy-specific phenotypic clusters that more accurately reflect the morphological landscape of the biopsy domain. Final cluster count \(c\) is reported in the Results section. All experiments were run with fixed random seeds and patient stratification to ensure reproducibility.

\begin{figure}[htbp]
\floatconts
  {fig:hpl}
  {\caption{\textbf{(a)} Overview of the analysis pipeline. From a dataset of 538,026 tiles, features were extracted using a frozen, pretrained (frozen) encoder to obtain embeddings. A random subsample of 250,000 embeddings was clustered using the Leiden algorithm, yielding 53 clusters. \textbf{(b)} Each cluster exhibits a distinct and internally consistent morphological pattern, illustrated by representative tile sets. The accompanying bar plot displays the number of tiles assigned to each cluster, highlighting the distribution of clusters within one fold of the cross-validation scheme.}}
  {\includegraphics[width=1\linewidth]{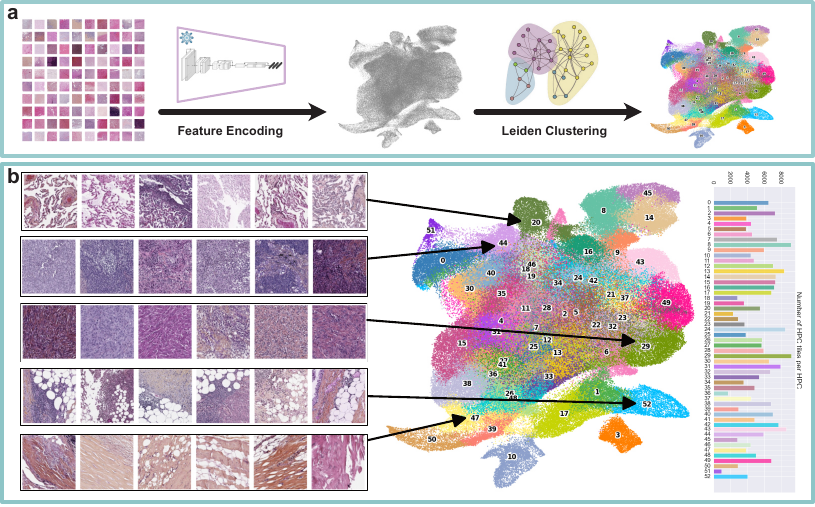}}
\end{figure}

\section{Results}
Using the Leiden algorithm, we identified 53 distinct HPCs in the biopsy dataset. The UMAP projection of tile embeddings is shown in Figure \ref{fig:hpl}b, with points coloured by cluster assignment. The feature encoder, trained exclusively on H\&E images, proved robust to colour variation, producing discriminative feature vectors that captured underlying morphological differences despite the presence of saffron staining. Representative tiles from different clusters highlight clear morphological patterns. For instance, cluster 20 exhibits open, lung-like architectures; clusters 44 and 29 display progressively denser cellular patterns; cluster 52 is enriched for adipose-like tissue; and cluster 47 corresponds to stroma-dominated regions, with collagen highlighted by the characteristic orange saffron stain. The adjacent bar plot shows the tile counts per cluster, illustrating the cluster distribution within a single fold of the cross-validation.

Patient-level phenotypic profiles were constructed by computing the proportional representation of each cluster within all patient's slides, and a Cox proportional hazards model was applied for survival prediction. The model achieved a C-index of 0.64 on the training set and 0.60 on the test set of the biopsy cohort (Table \ref{tab:performance}). Performance across datasets with varying exposure to the pretrained encoder is summarised in the benchmarking table. The encoder, trained exclusively on LATTICe-M, explains why both the Cox model and the epithelioid classifier perform best on this dataset. The TCGA-MESO cohort is substantially smaller and was not used for encoder training; its tiles were mapped to LATTICe-M clusters, resulting in lower classification performance. The final two rows correspond to train and test splits of the biopsy dataset, using clusters derived entirely from biopsy tissue. Despite this domain shift, the model generalises well, achieving a cross-validated AUC of 0.92 ± 0.03 with logistic regression and 0.90 ± 0.05 with the CLAM MIL approach on the test set.

\begin{table}[htbp]
\floatconts
  {tab:performance}%
  {\caption{ Benchmarking the model’s performance across multiple datasets. The model achieves performance on biopsy data comparable to its pretraining dataset (LATTICe-M) while also identifying novel clusters. Reported metrics are averaged across 5-fold cross-validation and are presented as mean ± standard deviation. }}%
  {\begin{tabular}{llll}
  \bfseries Dataset & \bfseries C-Index & \bfseries AUC (LR) & \bfseries AUC (CLAM)\\
  LATTICe-M & 0.65 ± 0.03& 0.88 ± 0.04 & 0.87 ± 0.04\\
  TCGA-Meso & 0.65 ± 0.01 & 0.8 ± 0.03& 0.74 ± 0.01 \\
  Biopsy (Train) & 0.64 ± 0.01 & 0.96 ± 0.01& 0.96 ± 0.01\\
  \textbf{Biopsy (Test)} & \textbf{0.6 ± 0.03} & \textbf{0.92 ± 0.03} & \textbf{0.9 ± 0.05}
  
  \end{tabular}}
\end{table}

Figure \ref{fig:downstream}a presents the Kaplan–Meier survival curves for the full biopsy cohort (n = 1062), combining both training and test patients from fold 0 after cross-validation. Patients were stratified into high- and low-risk groups based on their partial log-hazard values derived from the Cox model. Using the median hazard value as the cutoff resulted in a highly significant separation between the two groups (\(p = 3.96 \times 10^{-29}\)).

Analysis of the Cox model coefficients highlights cluster 0 as the strongest positive contributor to survival (\(p < 0.01\)). This cluster is enriched for adipose-associated inflammatory regions, potentially reflecting a more active host inflammatory response. In contrast, cluster 41 shows the most negative impact on survival, capturing aggressive, high-grade sarcomatoid and desmoplastic morphological patterns, well-established lethal phenotypes in mesothelioma. 

The logistic regression classifier, based on slide-level cluster compositions, accurately distinguished epithelioid from non-epithelioid slides in the biopsy test set, with an AUC-ROC of 0.92 ± 0.03, F1-score of 0.86 ± 0.06, recall of 0.82 ± 0.05, and precision of 0.92 ± 0.07. Analysis of the logistic regression coefficients reveals a strong inverse relationship between the predicted subtype and the morphological clusters. Cluster 12 serves as the strongest positive predictor for the non-epithelioid subtype, which is dominated by sarcomatoid-biphasic morphologies. Conversely, Cluster 25 exhibits the most negative coefficient, corresponding directly to the classic architecture of small infiltrative epithelioid cords and nests, as shown in Figure \ref{fig:downstream}b.

\begin{figure}[htbp]
\floatconts
  {fig:downstream}
  {\caption{Downstream WSI/Patient-level Tasks Performance. (a) Survival analysis results across the full cohort. The Cox proportional hazards model identifies several cluster-derived features with significant prognostic value (\(p < 0.01\)). Four representative tiles are shown for both high-risk and low-risk features to illustrate their underlying morphology. (b) Performance of the epithelioid vs. non-epithelioid classifier. The model achieves robust 5-fold cross-validated ROC–AUC, F1, recall, and precision on both training and test sets. Representative tiles from the clusters with the most negative and positive odds ratio and statistically significant contributions (\(p < 0.05\)) are also displayed.}}
  {\includegraphics[width=1\linewidth]{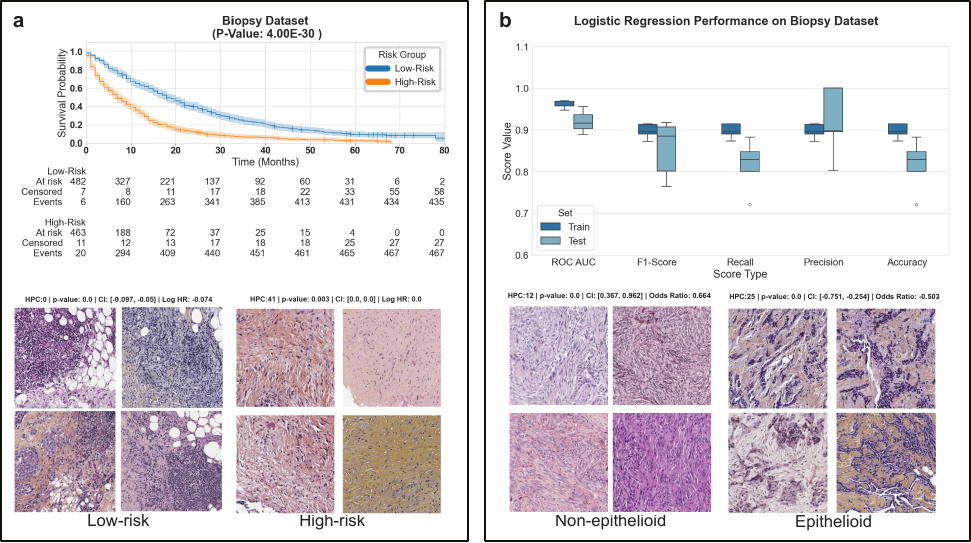}}
\end{figure}

\section{Discussion}
In this study, we evaluated the HPL-Meso encoder originally trained on H\&E surgical resections from the University Hospitals of Leicester, UK (LATTICe-M) on a heterogeneous cohort of HPS and HES biopsy slides collected across multiple histopathology centres in France. Our goal was to derive a new dictionary of biopsy-specific histomorphological phenotype clusters, and assess the utility of these features for downstream clinical tasks, including survival prediction and mesothelioma subtype classification, both of which have direct diagnostic and therapeutic impact in this disease.

The decision to reuse a mesothelioma-trained encoder, rather than a foundation model, was intentional. We aimed for a model specific enough to capture disease-relevant morphology while remaining generalisable across institutions and staining protocols. Training a bespoke model from scratch, particularly in clinical environments with limited computational resources, is rarely feasible, making pretrained encoders an essential component of practical digital pathology pipelines.

The key focus of this evaluation is the presence of two concurrent domain shifts: the application of a resection-trained model to biopsy material, and the transition from H\&E to HPS/HES staining; the latter introducing saffron to enhance collagen visualisation. Given that saffron-based stains are routinely used in France and French-speaking Canada but remain uncommon elsewhere, this setting provides a rigorous context in which to examine the model’s capacity to generalise across both tissue substrate and staining practice. Additionally, scanners and colour profiles differed substantially from those used to generate the original resection dataset. Despite this, the self-supervised encoder demonstrated strong colour invariance and successfully produced discriminative feature embeddings that organised biopsy tiles into well-separated phenotypic clusters. These clusters captured recognisable mesothelioma morphologies and were sufficiently stable to support downstream inference.

The clusters enabled meaningful survival modelling and delivered particularly strong performance in subtype classification (Table \ref{tab:performance}). The superior performance in the subtyping task (in compare to LATTICe-M) likely reflects the nature of biopsy tissue; biopsies concentrate on tumour-rich regions where possible, reducing morphological heterogeneity and making cellular architecture more diagnostic. Survival prediction, however, proved more difficult. Biopsies capture only a small portion of the tumour microenvironment, providing a limited biological snapshot for each patient. This constrained context reduces the prognostic information available to the model, which explains the lower performance compared with survival prediction on the original resection dataset.

Overall, our results show that models trained on resection specimens can still achieve clinically meaningful accuracy when applied to biopsy material. Notably, the biopsy validation also strengthens the evidence for the robustness of the HPL pipeline, even with a smaller cohort than LATTICe-M.

A limitation of this study is the incomplete transparency around the provenance of the external biopsy cohorts. Although we validated performance across two independent institutions and confirmed consistent results across the major scanner models used, we did not have access to fine-grained, site-specific metadata such as detailed instrument maintenance logs, reagent batch information, or full IT infrastructure records. This limits our ability to fully rule out subtle site-specific confounders and reflects an ongoing challenge in multi-institutional translational research. 

Another limitation is the subtype distribution within the external cohorts: our analysis primarily focused on distinguishing epithelioid from non-epithelioid cases, and while the model performed strongly, generalisation to the rarer biphasic and sarcomatoid subtypes remains constrained by limited sample sizes. Larger, multi-centre collaborations will be essential for rigorous evaluation of these rare entities. Furthermore, although we demonstrate that histology-derived phenotype features alone can support both diagnostic and prognostic tasks, the current framework does not integrate multimodal data such as next-generation sequencing or other genomic markers. Our survival analysis relied exclusively on morphological signals, and incorporating genomic features (such as BAP1 status) will be an important next step to quantify the added value of computational pathology features over established clinical and molecular diagnostics. This multimodal extension is already planned as a follow-up to the present work.

From a translational perspective, this validation framework offers a practical blueprint for integrating self-supervised pathology models into clinical workflows. It shows that pretrained, disease-specific encoders can be effectively repurposed for diagnostic stratification and risk assessment directly from routine biopsy slides, providing a scalable path toward AI-assisted mesothelioma diagnosis and clinical decision-making. While domain-specific differences inevitably introduce some uncertainty, systematic external validation on real-world biopsy data remains essential before clinical deployment.

To promote transparency and enable full reproducibility, we commit to releasing the biopsy-specific HPCs, upon request. The code for compositional feature extraction, and the trained SSL encoder weights are on public repository \cite{adalberto_claudio_quiros_2025_16947385}.

\clearpage  
\midlacknowledgments{We thank the Centre Léon Bérard (France) for their collaboration with Cancer Research UK and for providing the biopsy dataset, digital pathology infrastructure, and essential pathological expertise. We also acknowledge the Cancer Research UK Scotland Institute for their continued support of mesothelioma research programme.}

\bibliography{midl-samplebibliography}

\clearpage
\appendix
\section{Barlow Twins Training Formulation}
\label{barlow}
During training, two stochastic augmentations (views) \(v\) and \(v'\) are generated for each tile, producing embeddings \(z = f_{\theta}(v)\) and \(z' = f_{\theta}(v')\). For a mini-batch of size \(N\), the empirical cross-correlation matrix \(C\in\mathbb{R}^{D\times D}\) is defined as:
\begin{equation}
C_{ij} = \frac{1}{N}\sum_{b=1}^{N} \bar{z}_{b,i}\,\bar{z}'_{b,j},
\label{eq:crosscorr}
\end{equation}
where \(\bar{z}\) and \(\bar{z}'\) are the per-dimension mean-centred, variance-normalised embeddings across the batch. The Barlow Twins objective encourages invariance through diagonal alignment and decorrelation through off-diagonal suppression:
\begin{equation}
\mathcal{L}_{\text{BT}}(\theta)
= \sum_{i=1}^{D} (1 - C_{ii})^{2}
\;+\; \lambda \sum_{i=1}^{D}\sum_{j\neq i} C_{ij}^{2},
\label{eq:barlow}
\end{equation}
with \(\lambda>0\) controlling the off-diagonal penalty.

\paragraph{Inference.}
After training, each tile \(t\) is embedded via \(z = f_{\theta}(t)\) and used in the downstream HPL pipeline as described in the main text.

\section{Compositional Vectors}
\label{compositional}
We defined the raw HPC frequency vector for WSI \(s\) as
\begin{equation}
\tilde{\mathbf{a}}^{(s)} \;=\; \left( \tilde a^{(s)}_1, \dots, \tilde a^{(s)}_c \right),\qquad
\tilde a^{(s)}_j = \frac{n_{s,j}}{\sum_{u=1}^c n_{s,u}}.
\label{eq:rawfreq}
\end{equation}
By construction \(\tilde{\mathbf{a}}^{(s)}\) is compositional: \(\tilde a^{(s)}_j\ge 0\) and \(\sum_j \tilde a^{(s)}_j=1\).

These vectors can contain zeros; we apply multiplicative replacement prior to log-ratio transformations. We then transform each compositional vector \(\mathbf{a}^{(s)}\) into Euclidean space via the centered log-ratio:
\begin{equation}
\mathrm{clr}\!\left(\mathbf{a}^{(s)}\right) \;=\; \left( \log\frac{a^{(s)}_1}{g(\mathbf{a}^{(s)})},\ \dots,\ \log\frac{a^{(s)}_c}{g(\mathbf{a}^{(s)})} \right)
\label{eq:clr}
\end{equation}
where \(g(\mathbf{a}) = \big(\prod_{j=1}^c a_j\big)^{1/c}\) is the geometric mean. The clr vector has a zero sum: \(\sum_j \mathrm{clr}(\mathbf{a})_j = 0\).





\end{document}